\documentclass[10pt]{asme2ej}

\usepackage{epsfig} 
\usepackage{subfigure}
\usepackage{url}
\usepackage{amsmath}

\title{Multifractal Terrain Generation for Evaluating Autonomous Off-Road Ground Vehicles}

\author{Casey D. Majhor\thanks{Corresponding author.}
    \affiliation{
	Robust Autonomous Systems Laboratory\\
	Department of Electrical and Computer Engineering\\
	Michigan Technological University\\
	Houghton, Michigan 49931\\
    Email: cmajhor@mtu.edu
    }	
}

\author{Jeremy P. Bos
    \affiliation{
	Robust Autonomous Systems Laboratory\\
	Department of Electrical and Computer Engineering\\
	Michigan Technological University\\
	Houghton, Michigan 49931\\
    Email: jpbos@mtu.edu
    }
}

\begin{document}

\maketitle    

\begin{abstract}
{\it 
We present a multifractal artificial terrain generation method that uses the 3D Weierstrass-Mandelbrot function to control roughness. By varying the fractal dimension used in terrain generation across three different values, we generate $60$ unique off-road terrains. We use gradient maps to categorize the roughness of each terrain, consisting of low-, semi-, and high-roughness areas. To test how the fractal dimension affects the difficulty of vehicle traversals, we measure the success rates, vertical accelerations, pitch and roll rates, and traversal times of an autonomous ground vehicle traversing $20$ randomized straight-line paths in each terrain. As we increase the fractal dimension from $2.3$ to $2.45$ and from $2.45$ to $2.6$, we find that the median area of low-roughness terrain decreases $13.8\%$ and $7.16\%$, the median area of semi-rough terrain increases $11.7\%$ and $5.63\%$, and the median area of high-roughness terrain increases $1.54\%$ and $3.33\%$, all respectively. We find that the median success rate of the vehicle decreases $22.5\%$ and $25\%$ as the fractal dimension increases from $2.3$ to $2.45$ and from $2.45$ to $2.6$, respectively. Successful traversal results show that the median root-mean-squared vertical accelerations, median root-mean-squared pitch and roll rates, and median traversal times all increase with the fractal dimension.
}
\end{abstract}

\section{Introduction}
Off-road vehicle autonomy poses unique challenges compared to their on-road counterparts~\cite{carruth2022challenges}. These challenges include operating in rough terrain, and without designated roads which typically outline safe and reasonable paths to traverse. Furthermore, traversing terrain beyond an autonomous ground vehicle's (AGV's) capabilities can cause immobilizing damage~\cite{darpa2004,thrun2006stanley}.

Testing AGVs in varied terrains is crucial to evaluate their performance. However, field testing vehicles can be expensive, time-consuming, and risky. As an alternative, software-in-the-loop has become a useful tool. Software-in-the-loop allows for quick and repeated AGV testing in different scenarios. Furthermore, it is important for AGV researchers to have control over terrain roughness. Testing in diverse terrains prevents biased results skewed to a particular terrain type. Thus, a method to generate terrain with roughness control for AGV simulations is desirable.

Procedural approaches are commonly used to generate terrain using algorithms with minimal user input~\cite{smelik2014survey}. These algorithms are often based on fractals or textural noise~\cite{mandelbrot1975stochastic, gasch2020procedural}, geological simulation such as erosion~\cite{doran2010controlled, genevaux2013terrain}, or data-driven approaches~\cite{zhou2007terrain, zhang2022authoring}. However, relatively few studies focus on terrain generation for off-road AGV testing. Related work in Refs.~\cite{dawkins2012fractal,dawkins2012evaluation,dawkins2014model}, applied the 3D Weierstrass-Mandelbrot (W-M) fractal function for terrain profile generation for simulating off-road vehicle dynamics.

Fractal-based methods such as the 3D W-M function are a popular approach for terrain and surface generation. These methods incorporate random parameters into their respective mathematical functions. Because of their stochastic nature, these methods have been shown to work well for generating unique terrains~\cite{ausloos1985multivariate, dawkins2014model}. However, fixed parameters used in these algorithms, which define the terrain elevations, tend to result in stationary and isotropic terrain, meaning that spatially their statistical properties do not change significantly~\cite{dawkins2012evaluation}. When using the W-M function, a single fractal dimension ($D$) parameter creates monofractal terrain which tends to be too stationary and isotropic for testing AGV path planning and navigation methods.

\begin{figure*}[hbt!]
\centering
\subfigure[]{\label{fig:UE_base2_low_top}\includegraphics[width=1.68in]{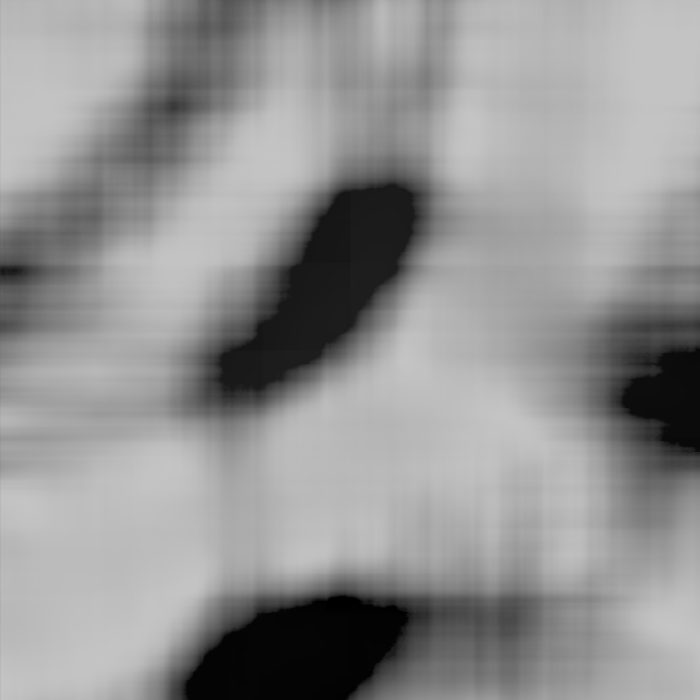}}
\subfigure[]{\label{fig:UE_base2_mid_top}\includegraphics[width=1.68in]{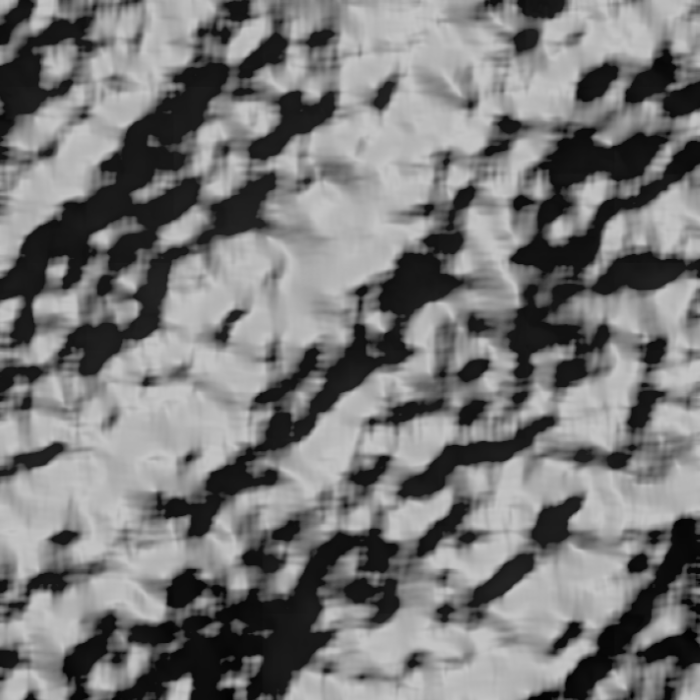}}
\subfigure[]{\label{fig:UE_base2_high_top}\includegraphics[width=1.68in]{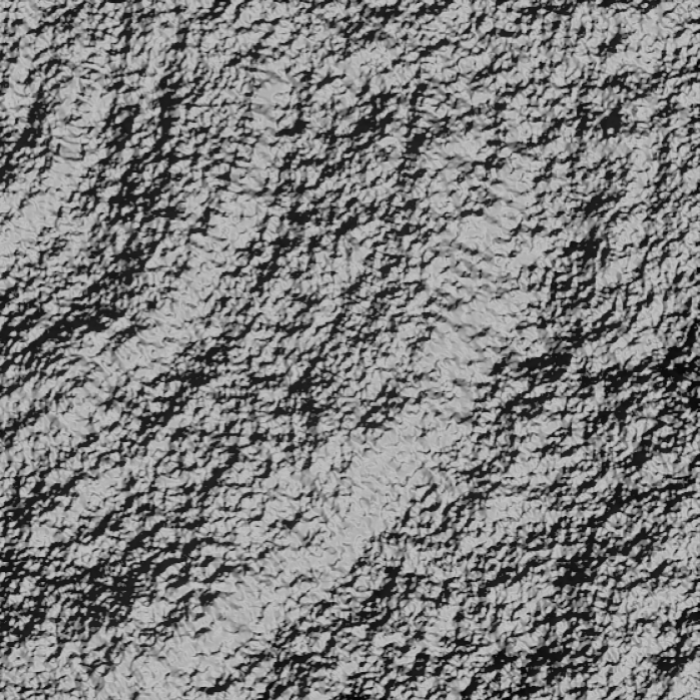}}
\subfigure[]{\label{fig:UE_base2_comb_top}\includegraphics[width=1.68in]{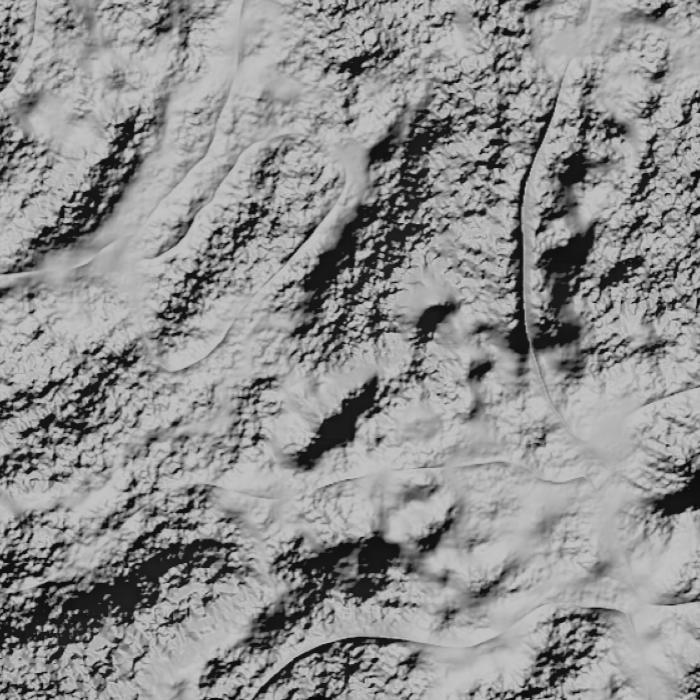}}
\subfigure[]{\label{fig:UE_base2_low_perspective}\includegraphics[width=1.68in]{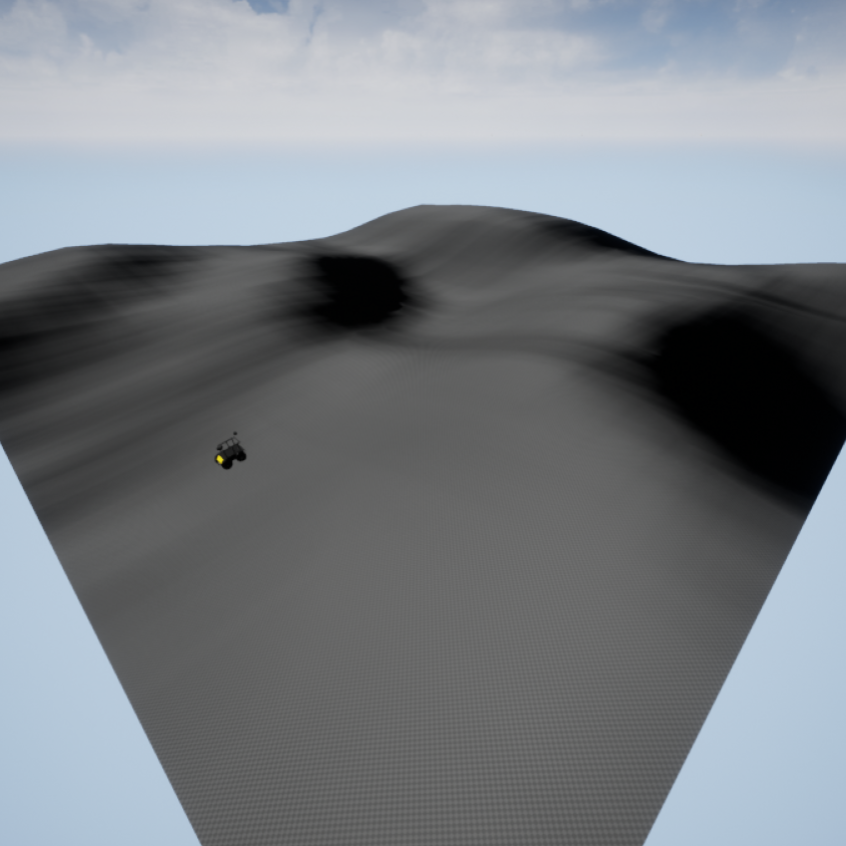}}
\subfigure[]{\label{fig:UE_base2_mid_perspective}\includegraphics[width=1.68in]{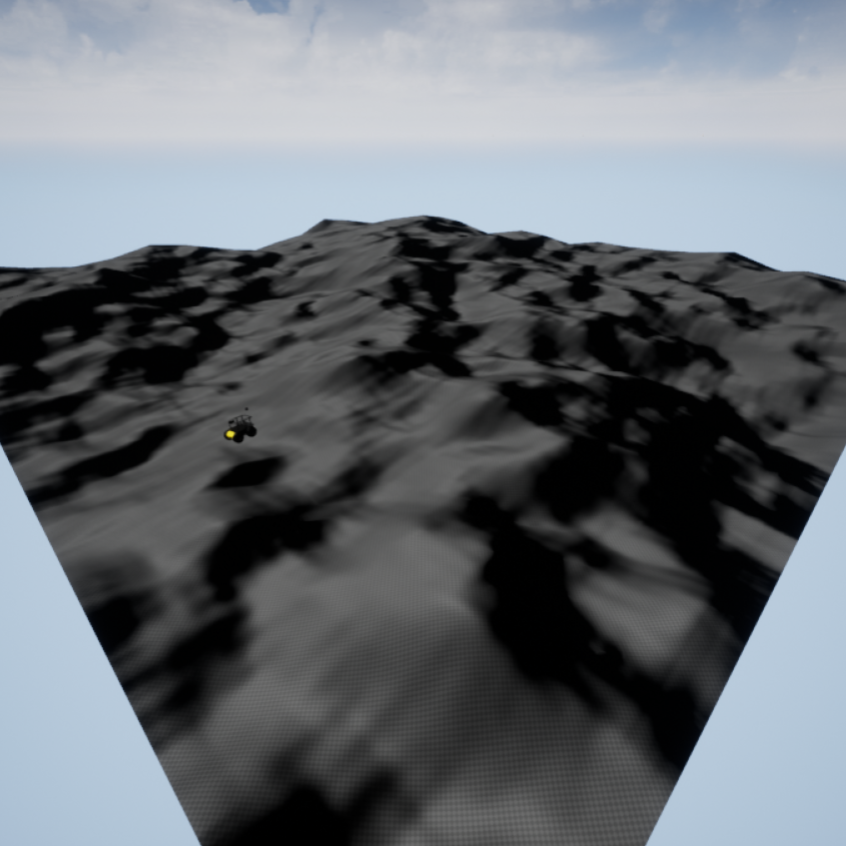}}
\subfigure[]{\label{fig:UE_base2_high_perspective}\includegraphics[width=1.68in]{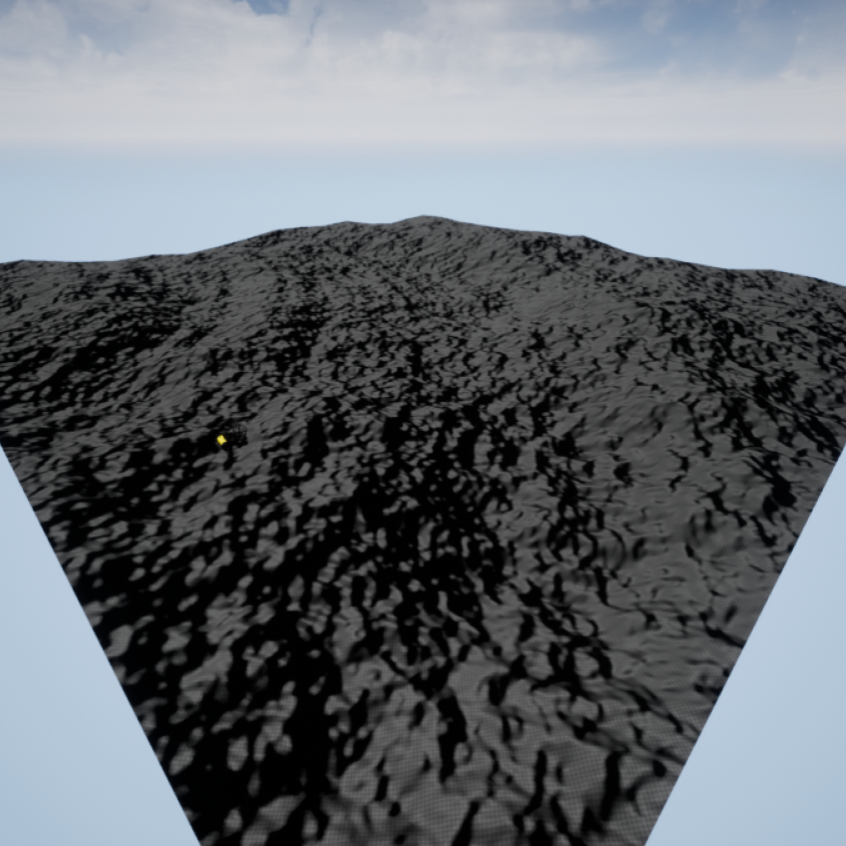}}
\subfigure[]{\label{fig:UE_base2_comb_perspective}\includegraphics[width=1.68in]{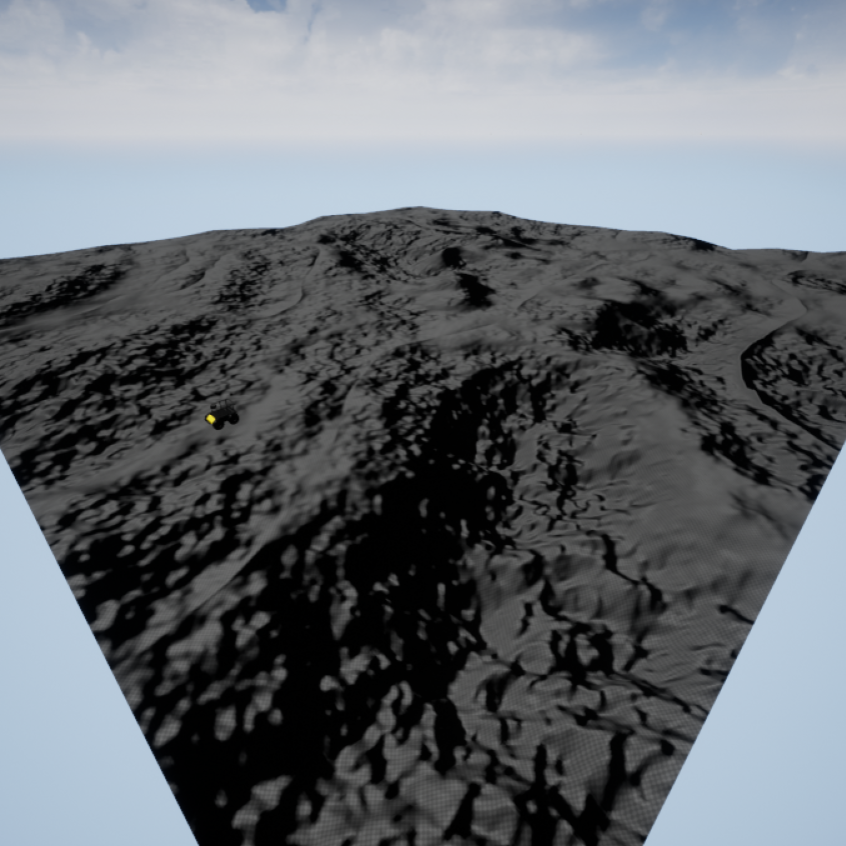}}
\caption{Top and perspective views of three example monofractal terrains (a, c), (b, f), (c, g) generated using the 3D Weierstrass-Mandelbrot function using different parameters. An example multifractal terrain (d,h) using our approach which combines the three example terrains to create more varied roughness for autonomous ground vehicle testing.}
\label{fig:UE_low_mid_high_comb_views}
\end{figure*}

While monofractal terrains have useful applications~\cite{dawkins2014model}, they are trivial for testing off-road AGV path planning and navigation algorithms since they tend to either be too smooth or too rough. As an example, Figs.~\ref{fig:UE_base2_low_top}-\ref{fig:UE_base2_high_perspective} shows top and perspective views of three monofractal terrains generated by the 3D W-M function with fixed fractal dimensions. A modeled Clearpath Husky AGV used in this work is shown in perspective views for scale. Each of these three terrains shows little variation in roughness. As we demonstrate in this work, more variable roughness can be achieved using the W-M function by using multiple fractal dimensions, resulting in multifractal terrain.

In this work, to overcome the shortcomings of monofractals for AGV testing, we combine monofractal surfaces to create multifractal terrain. Our multifractal terrain generation creates terrain with variable roughness such that it contains both traversable and untraversable segments. Path planning and navigating through multifractal terrain with varied roughness requires specific terrain features to be avoided to ensure A-to-B traversal mission success. Hence, testing AGVs for path planning and navigation in varied terrains will yield more meaningful results. An example of multifractal terrain generated using our approach is shown in Figs.~\ref{fig:UE_base2_comb_top} and \ref{fig:UE_base2_comb_perspective}, achieved by combining the three terrains in Figs.~\ref{fig:UE_base2_low_top}-\ref{fig:UE_base2_high_perspective}.

In this paper, we generate $50.4\times50.4~m$ terrain surfaces as opposed to the tire-width terrain profiles generated in Refs.~\cite{dawkins2012fractal,dawkins2012evaluation,dawkins2014model}. Three-dimensional terrain surfaces are required to test AGV planning and navigation algorithms such that the AGV is not restricted to a single direction of motion. The terrain profiles in Refs.~\cite{dawkins2012fractal,dawkins2012evaluation,dawkins2014model} do not allow for simulating vehicle turns or lateral movements which are important to consider in path planning and navigation. Furthermore, our terrain is more anisotropic in roughness which is desirable so AGV path planning and navigation scenarios are nontrivial. The terrain generation methods in Refs.~\cite{vecchio2022midgard,hudson2020mississippi} are not readily adaptable to other simulators since they are each integrated with specific simulation environments for full-scene generation. Other approaches such as in Ref.~\cite{toupet2020ros} are tailored towards specific terrain types, Martian terrain in this case, which can limit its use. Our terrain generation method produces digital elevation maps (DEMs), which can easily be adapted to other simulation environments~\cite{collins2021review} such as Gazebo~\cite{koenig2004design}, CoppeliaSim (V-REP)~\cite{CoppeliaSim}, and Webots~\cite{michel2004cyberbotics}.

Our terrain generation approach utilizes the 3D W-M function to generate multifractal terrains. To test if terrain roughness can be controlled, we vary the fractal dimension used in terrain generation across three different values, generating $60$ unique DEMs. We use gradient maps to categorize the composition of each DEM, consisting of low-, semi-, or high-roughness areas. To test if and how $D$ affects the difficulty of vehicle traversals, the DEMs are rendered in Unreal Engine (UE)~\cite{unrealengine_misc} for AGV simulations. We measure the success rates, vertical accelerations, pitch and roll rates, and traversal times of an AGV traversing $20$ straight-line paths with randomized start and goal locations in each terrain. 

As we increase the fractal dimension from $2.3$ to $2.45$ and from $2.45$ to $2.6$, we find that the median area of low-roughness terrain decreases $13.8\%$ and $7.16\%$, the median area of semi-rough terrain increases $11.7\%$ and $5.63\%$, and the median area of high-roughness terrain increases $1.54\%$ and $3.33\%$, all respectively. We find that the median success rate of the vehicle decreases $22.5\%$ and $25\%$ as the fractal dimension increases from $2.3$ to $2.45$ and from $2.45$ to $2.6$, respectively. Successful AGV trials show that the median root-mean-squared (RMS) vertical accelerations, median RMS pitch and roll rates, and median traversal times increase with $D$ as well. 

We show that we can control terrain roughness and affect traversal performance. Our approach allows for measuring AGV performance over trials where the surface characteristics are varied. If AGV path planning and navigation algorithms are only tested on terrains with similar surface characteristics, the results will only apply to those terrain characteristics. As we will show, terrains produced by our approach contain varying amounts of low-, semi-, and high-roughness areas. By testing AGVs in terrains with varied surface characteristics, we can avoid results that are biased towards specific terrains. Control over the terrain roughness with our approach allows AGVs to be easily tested in various levels of terrain difficulty. 

The rest of this paper is structured as follows. We begin by outlining our methods in Sec.~\ref{sec:methodology}. This methodology section describes our terrain generation approach, A-to-B random mission selection, experiments, and evaluation metrics. We then present the experimental results and discussion in Sec.~\ref{sec:results}. Finally, conclusions and future work are presented in Sec.~\ref{sec:conclusions}.

\section{Methodology} 
\label{sec:methodology}
In this section, we first outline our proposed multifractal off-road DEM generation method. Next, the process for rendering the DEMs in UE is explained. Then the DEM gradient map filtering and random mission selection processes are presented. We then outline the AGV traversal experiments and the evaluation metrics used.

\subsection{Multifractal DEM Generation}
\label{sec:proc_gen_terrain}
To generate our terrain, we use a fractal technique adopted from Dawkins et al.~\cite{dawkins2014model}. We assign elevation values ($z$) to our DEMs at gridded $x$ and $y$ locations using the 3D W-M function~\cite{ausloos1985multivariate}. This function is defined as
\begin{equation}
z(x, y) = A \sum_{m=1}^{M} \sum_{n=1}^{n_{\text{max}}} \gamma^{(D-3)n} \left[ \cos(\phi_{m,n}) - \cos \left\{ \frac{2\pi \gamma^{n} \sqrt{x^{2} + y^{2}}}{L} \cos \left( \arctan \left( \frac{y}{x} \right) - \frac{\pi m}{M} \right) + \phi_{m,n} \right\} \right],
\label{eq_WM}
\end{equation}
where
\begin{equation}
A = L\left(\frac{G}{L}\right)^{D-2}\left(\frac{ln \left(\gamma\right)}{M}\right)^{1/2}.
\label{eq_A}
\end{equation}

The W-M function operates as a summation of $M$ surface layers of ridge-like structures. Thus, $M$ is the number of ridges in Eqs.~\ref{eq_WM}-\ref{eq_A}. The $n$ parameter is a frequency index, with $n_{max}$ being an upper cutoff frequency. We choose $n_{max}$ to be the length of our square DEMs in pixels. The $\gamma$ parameter controls the density of surface frequencies, and $D$ is the fractal dimension. The $\phi$ values are uniform random phases among the sinusoids. These random phases allow for unique DEMs to be generated using different random seeds. The parameter $L$ is a sampling length, and $G$ is an elevation scaling coefficient. 

\begin{figure*}[hbt!]
\centering
\subfigure[]{\label{fig:base2_low_freq_mesh}\includegraphics[width=1.68in]{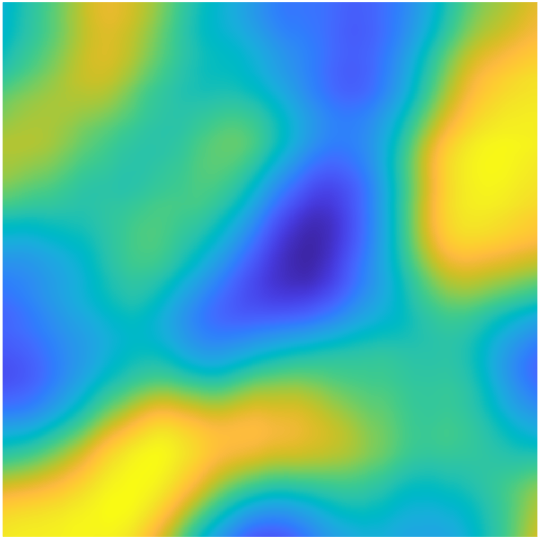}}
\subfigure[]{\label{fig:base2_mid_freq_mesh}\includegraphics[width=1.68in]{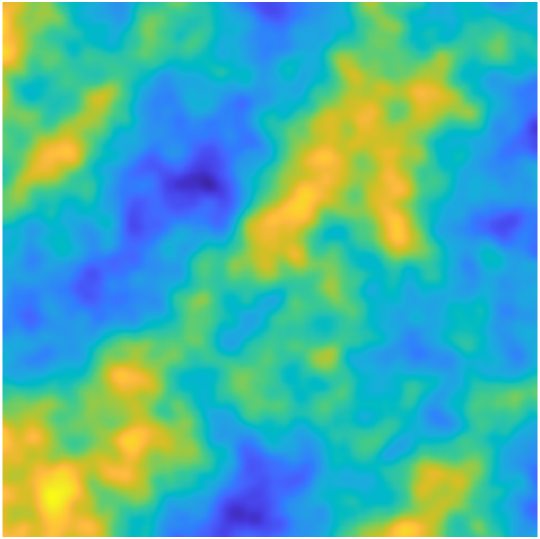}}
\subfigure[]{\label{fig:base2_high_freq_mesh}\includegraphics[width=1.68in]{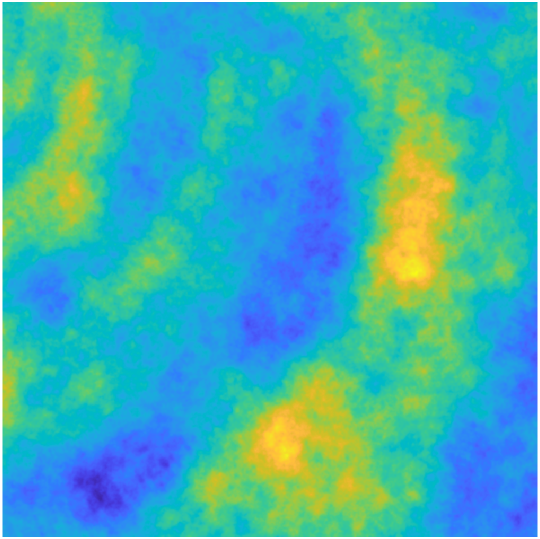}}
\subfigure[]{\label{fig:base2_NewSurface_DEM}\includegraphics[width=1.68in]{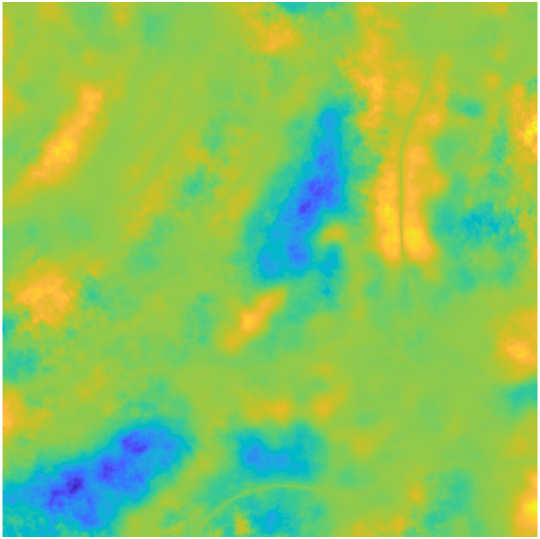}}
\caption{Example low- (a), mid- (b), and high-frequency (c) monofractal digital elevation maps (DEMs), combined via the pixel-wise product to create a fourth multifractal DEM (d) which is imported into Unreal Engine for autonomous ground vehicle testing. Note that (a-c) have Weierstrass-Mandelbrot ridges ($M$) equal to 16, 32, and 64, fractal dimensions ($D$) equal to 2.2, 2.45, and 2.45, and elevation scaling coefficients ($G$) equal to 1e-6, 8e-8, and 1e-8, all respectively.}
\label{fig:sub_dems}
\end{figure*}

To create more anisotropic terrain, we use an integrative approach that combines three DEMs, each with different scaling and orientations of elevation features. To achieve this, we use the W-M function and vary the number of ridges ($M$), fractal dimension ($D$), and elevation scaling coefficient ($G$). Each of the three DEMs uses an $M$ value of either $16$, $32$, or $64$, determining their spatial frequency. We refer to these as the low-, mid-, and high-frequency DEMs, respectively. To smooth the DEMs, we use a low-pass Gaussian filter. The three DEMs are all scaled near the same range and with zero mean. This scaling negates the decreased DEM scaling apparent when $D$ increases, defined in Eq.~\ref{eq_A}. The three DEMs are multiplied with the pixel-wise product to combine their features into a multifractal DEM. Figure~\ref{fig:sub_dems} shows an example of low-~(\ref{fig:base2_low_freq_mesh}), mid-~(\ref{fig:base2_mid_freq_mesh}), and high-frequency~(\ref{fig:base2_high_freq_mesh}) DEMs and their combined multifractal DEM~(\ref{fig:base2_NewSurface_DEM}). The multifractal DEM's features and roughness vary more by the spatial scale and are anisotropic compared to its comprised three DEMs, as shown in Fig.~\ref{fig:base2_NewSurface_DEM}. To meet UE's DEM import requirements, we convert the multifractal DEM into a grayscale, $16$-bit PNG image. Note that the four DEMs in Fig.~\ref{fig:sub_dems} correspond to the rendered terrains in Fig.~\ref{fig:UE_low_mid_high_comb_views}. An example DEM grayscale image is shown in Fig.~\ref{fig:multi_fract_matlab}. The multifractal DEMs are then rendered as a terrain in UE for AGV testing, as described in Sec.~\ref{sec:terrain_rendering}.

To test if we can control terrain roughness with our method, we vary $D$ when creating the high-frequency DEM. To limit confounding variables, and because terrain roughness is found to be interdependent on both $G$ and $D$~\cite{dawkins2012fractal}, only $D$ is adjusted in the high-frequency DEM. We modify $D$ only in the high-frequency DEM because we anticipate this to have the most impact on roughness at smaller spatial scales relevant to the size of the AGV we are testing. Adjusting $D$ has been found to affect near-isotropic roughness in terrain profiles for AGV testing in prior work~\cite{dawkins2012evaluation}. We reference this prior work for selecting the W-M function parameters. However, the effects of these parameters on our approach to creating more anisotropic terrain surfaces are unknown. To assess the impact of $D$ on terrain roughness, we select $D$ values of $2.3$, $2.45$, and $2.6$. These values sample $D$ between $2$ and $3$, commonly done in prior works. This sampling range preserves the fractal properties of homogeneity and scaling.~\cite{ausloos1985multivariate}. We use these values to test whether we can control the resulting roughness of the terrain. The W-M function parameters defining the low-, mid-, and high-frequency DEMs in our study are found in Table~\ref{tab:WM_params}.

\begin{table}[h!]
\caption{3D Weierstrass-Mandelbrot function parameters for the three monofractal digital elevation maps (DEMs) combined to create a fourth multifractal DEM with variable roughness. Note the boldface fractal dimension ($D$) parameters in the high-frequency DEM are adjusted to demonstrate control of terrain roughness and a correlation between $D$ and vehicle traversability.}
\begin{center}
\label{tab:WM_params}
\begin{tabular}{l | c c c}
\hline
\hline
Parameter & \multicolumn{3}{c}{Digital Elevation Map} \\
           & Low Frequency & Mid Frequency & High Frequency\\
\hline
$M$ &  $16$ & $32$ & $64$ \\ \hline
$n_{max}$ & $1009$ & $1009$ & $1009$\\ \hline
$\gamma$ & $1.5$ & $1.5$ & $1.5$\\ \hline
$D$ &  $2.2$ & $2.45$ & \textbf{2.3}, \\ 
  &        &        & \textbf{2.45}, \\ 
  &        &        & \textbf{2.6}\\ \hline
$\phi$ (random) & [0-$\pi$] & [0-$\pi$] & [0-$\pi$]\\ \hline
$L$ &  $100.9$ & $100.9$ & $100.9$\\ \hline
$G$ &  $1\times10^{-6}$ & $8\times10^{-8}$ & $1\times10^{-8}$\\
\hline
\hline
\end{tabular}
\end{center}
\end{table}

\subsection{Terrain Rendering in Unreal Engine}
\label{sec:terrain_rendering}
To test how $D$ affects AGV traversal difficulty, we simulate AGV traversals on our generated terrain by adapting work by Young et al.~\cite{young2020unreal}. Young et al. developed a simulation environment compatible with Robot Operating System (ROS)~\cite{ros_misc}. The authors use UE which includes a simulated Clearpath Husky AGV~\cite{husky}. The physics engine subsystem used in UE is NVIDIA's PhysX~\cite{physx}. PhysX accurately models vehicle-terrain interaction~\cite{erez2015simulation}. To plan and track straight-line paths with the AGV, we adopt a short-range trajectory planner and pure-pursuit path tracker from Refs.~\cite{krusi2017driving,kysar2021unstructured}.

\begin{figure}[hbt!]
\centering
\subfigure[]{\label{fig:multi_fract_matlab}\includegraphics[width=1.68in]{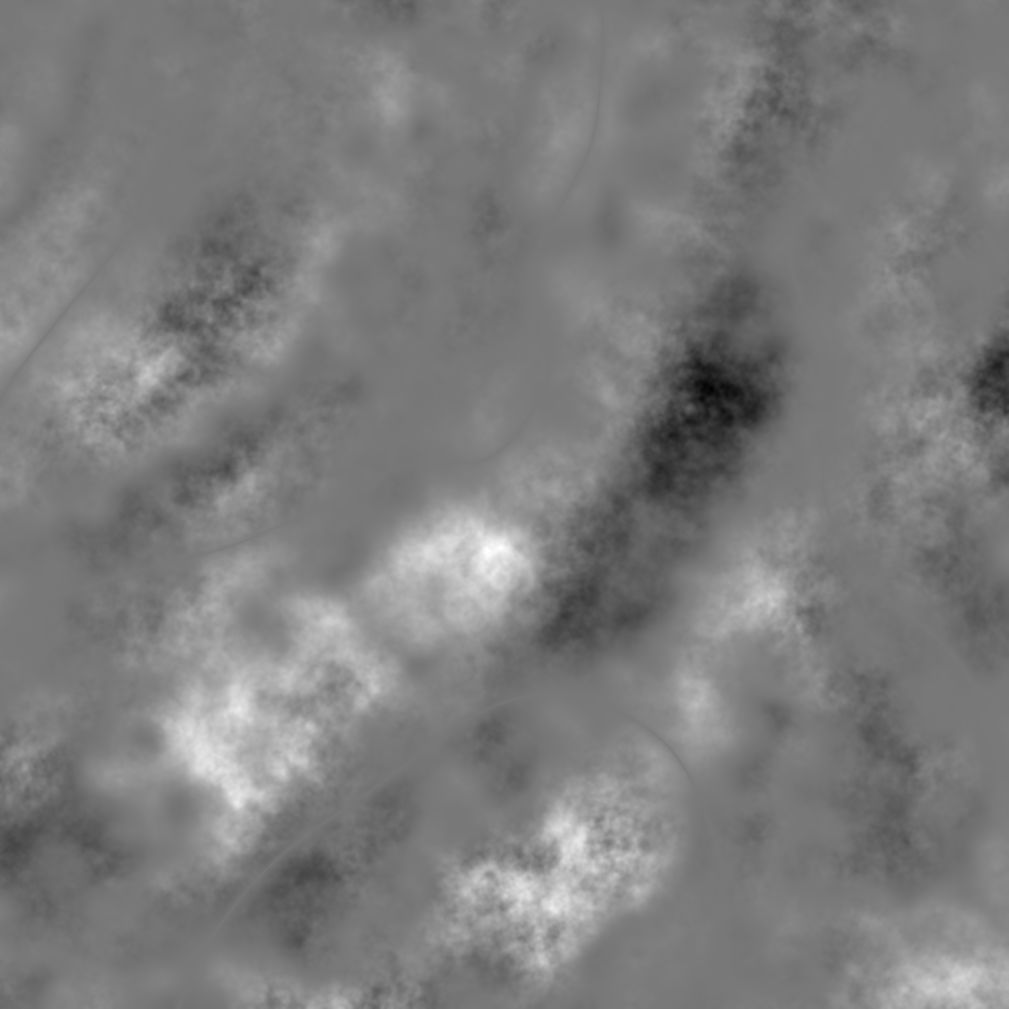}}
\subfigure[]{\label{fig:UE4_terrain_bird}\includegraphics[width=1.68in]{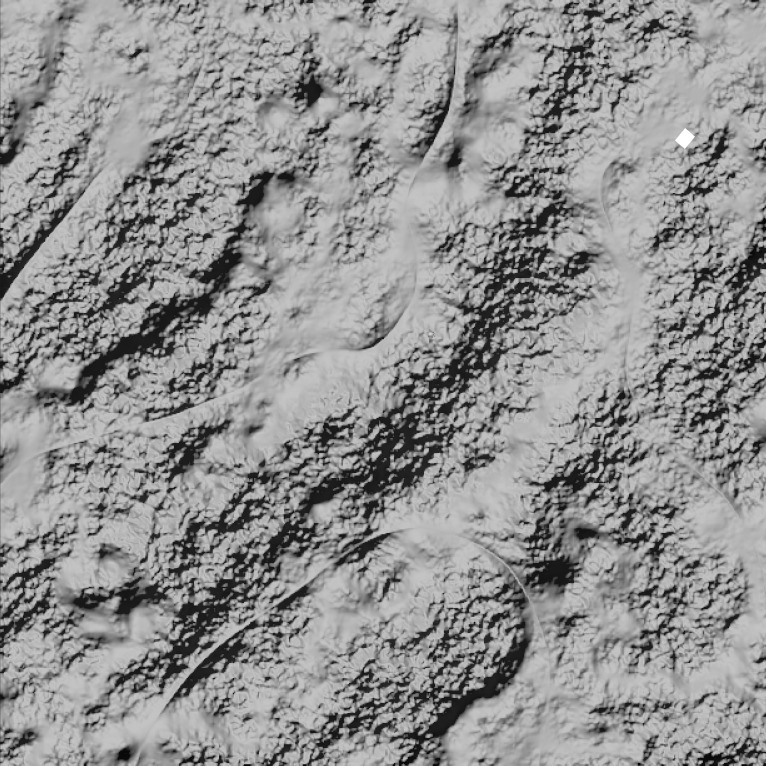}}
\subfigure[]{\label{fig:UE4_terrain_perspective}\includegraphics[width=3.36in]{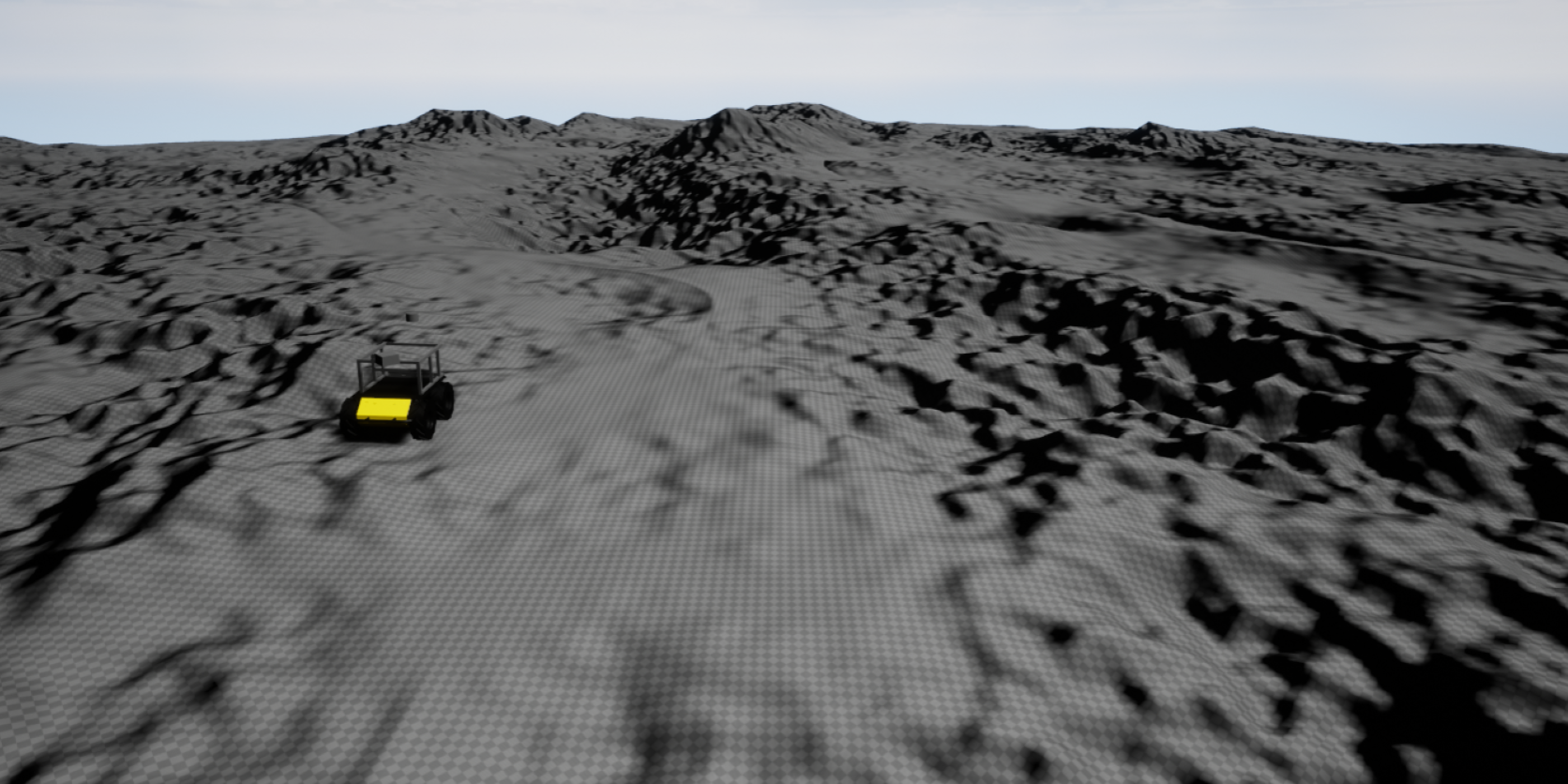}}
\caption{An example multifractal digital elevation map (DEM) with a top-down view of its grayscale image (a). The DEM in (a) is imported, scaled, and rendered in Unreal Engine (b). A perspective view of the Unreal Engine-rendered terrain is shown in (c) with a simulated Clearpath Husky autonomous ground vehicle in view, with its corresponding map location shown in (b) as a white rectangle.}
\end{figure}

After combining the low-, mid-, and high-frequency DEMs into a multifractal DEM, we render it in UE as a 3D terrain. The DEM is first encoded as a $1009\times1009$ pixel PNG and then is imported into UE~\cite{UE_heightmap}. Once imported, we scale the imported UE landscapes down by $95\%$ in the X and Y directions. This results in a resolution of $5~cm$ per pixel, and $50.4\times50.4~m$ terrain maps. The default terrain Z (elevation) scaling in UE of $100\%$ was found to be too large to create sections of immobilizing and smooth terrain. We experimented with the AGV in terrains generated with a $D$ value of $2.45$ to choose an appropriate terrain Z scaling. We found a Z scaling of $0.75\%$ appropriate such that the terrain contained areas that were both smooth and rough. An example 3D rendered terrain in UE, with top-down and perspective views, is shown in Figs.~\ref{fig:UE4_terrain_bird}, and \ref{fig:UE4_terrain_perspective}, respectively.

\subsection{DEM Filtering and Random Path Selection}
\label{sec:dem_filt_rand_sel}
We use Monte Carlo simulation to see if $D$ correlates with AGV traversability. We program the AGV to traverse $20$ random A-to-B straight-line-path missions at a speed of $1~m/s$, in $60$ different terrain maps. Since the terrains are obstacle-free, the straight-line paths represent a naive, non-terrain-aware path planning and navigation method. To ensure the missions are practical, we restrict the AGV from starting or ending paths in unreasonably rough terrain. We classify terrain as unreasonably rough by calculating the Moore neighborhood~\cite{biswas2018robust, grad8} gradients with a $3\times3$ pixel mask. This mask size is appropriate given that our DEM resolution is $5~cm$ per pixel, and the AGV's tire radius and width are $16.5$ and $12.5~cm$, respectively. The maximum absolute value of the neighborhood gradients is calculated for each DEM pixel, described as  
\begin{equation}
P_{grad} = max(abs(\{Q_{grad}\} ))
\label{eq_grad1}
\end{equation}
where
\begin{equation}
\left\{Q_{grad}\right\} = (Z_P - Z_{Q_{i}})/dist(P,Q_i).
\label{eq_grad2}
\end{equation}

In Eq.~\ref{eq_grad1}, $Q_{grad}$ denotes the set of gradient values in the Moore neighborhood of a center pixel \emph{P}. The functions $max()$ and $abs()$ represent the maximum and absolute value functions, respectively. $P_{grad}$ represents the maximum absolute valued gradient of a target (center) pixel. The Moore neighbors of pixel $P$ share a vertex or edge with it. Equation~\ref{eq_grad2} shows how the gradient set calculation is done. Here, $Z_P$ and $Z_{Q_{i}}$ are the DEM elevation values of the center pixel $P$ and neighbor pixel $Q_i$. The Euclidean distance between the two pixels is calculated and represented by the $dist()$ function. This distance equals $5$ and $7.07~cm$ for neighbors with a shared edge and vertex, respectively.

\begin{figure}[hbt!]
\centering
\subfigure[]{\label{fig:base2_grad_map}\includegraphics[width=1.68in]{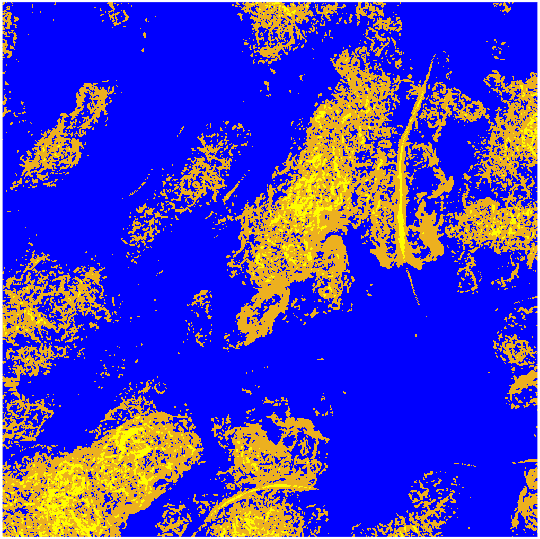}}
\subfigure[]{\label{fig:base2_grad_map_filt}\includegraphics[width=1.68in]{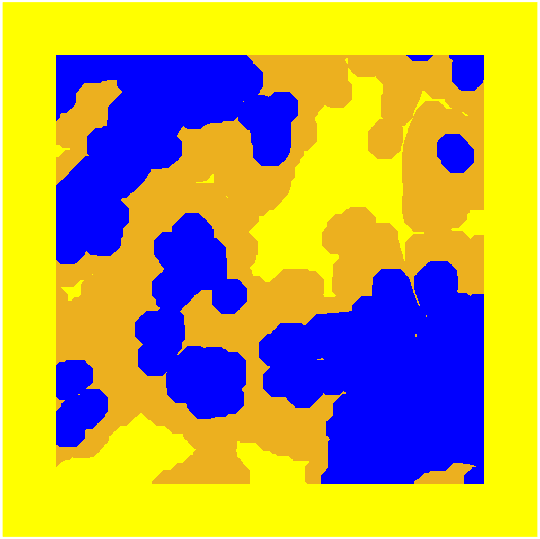}}
\subfigure[]{\label{fig:base2_grad_map_filt_rand}\includegraphics[width=1.68in]{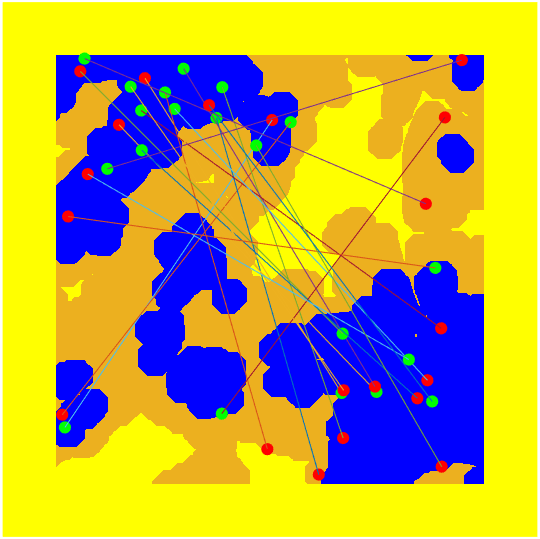}}
\caption{An example Moore-neighborhood-gradient map with pixels categorized into one of three groups: low (blue), semi (orange), or high roughness (yellow) (a). The gradient map from (a) is then filtered with a morphological closing filter for random mission selection, where high-roughness areas cannot contain a start/goal location, semi-rough areas cannot contain a start location, and low-roughness areas are free to contain start/goal locations (b). The filtered gradient map from (b) is shown with 20 randomly selected missions with straight lines connecting respective start (green) and goal (red) locations (c).}
\end{figure}

Equation~\ref{eq_grad1} is applied to all the DEM pixels, creating a gradient map. We adapt this approach for mission selection and terrain analysis. Each pixel in the gradient map is classified as having low, semi, or high roughness. Gradient values, as calculated in Eq~\ref{eq_grad2}, are expressed as unitless rise-over-run values. Pixels with a maximum gradient less than or equal to $50$ are classified as having low roughness. Semi-rough classified pixels have a maximum gradient greater than $50$ and less than or equal to $140$. Pixels with a maximum gradient above $140$ are classified as having high roughness. These category thresholds were determined by deploying the AGV and traversing UE terrain locations corresponding to the DEM gradient map. We found low-roughness terrain to be smooth and easily navigable. Terrain areas classified as semi-rough were moderately rough visually, but less navigable than low-roughness terrain. We found high-roughness areas very rough visually and likely to cause the AGV to become stuck or rollover. We define these categories to quantify the areas of roughness for each DEM, which will be compared with the respective $D$ values used. An example gradient map is presented in Fig.~\ref{fig:base2_grad_map}. Low-, semi-, and high-roughness areas are color-coded as blue, orange, and yellow, respectively.

We apply a morphological closing filter to the gradient maps to enclose rougher areas. For this, a disk structuring element is used, with effects shown in Fig.~\ref{fig:base2_grad_map_filt}. We also restrict start/goal locations not to be within five meters of the map edge. This start/goal restriction is shown as a yellow border in Fig.~\ref{fig:base2_grad_map_filt}. Additionally, we enforce starting locations to be in the low-roughness (blue) category. These restrictions help prevent the AGV from rolling from the deployment location before the mission starts. The restrictions also prevent the AGV from falling off the landscape in simulation. Goal locations must be in either the low or semi-rough (orange-coded) category so the A-to-B paths are more practical.

To test terrain traversability, we conduct randomized straight-line path missions with the AGV. The start and goal poses are separated by a Euclidean distance of $35~m$. Given our operational area with safety borders removed is $40.4\times40.4~m$, the $35~m$ missions can be in any orientation, given the start and goal satisfy the roughness criteria. This mission length also provides an appropriate sampling of the terrain. We select a mission starting pixel randomly from the low-roughness categorized indices. We set the starting pose by randomly selecting a heading between $0$ and 360\textdegree. The candidate goal pose, if valid, lies along the heading $35~m$ away. Figure~\ref{fig:base2_grad_map_filt_rand} shows $20$ example missions with start (green) and goal (red) locations. The path to traverse is shown with a line connecting each mission start and goal. Note that most of the missions in Fig.~\ref{fig:base2_grad_map_filt_rand} include all three roughness types.

\subsection{Experiments and Evaluation Metrics}
\label{sec:experiments}
In this work, we adjust the fractal dimension ($D$) to control roughness for our terrain generation method. We generate $20$ unique DEMs for each of the three $D$ values as described in Sec.~\ref{sec:proc_gen_terrain}. We examine how increasing $D$ from $2.3$ to $2.45$ and $2.6$ in the high-frequency DEM affects the roughness of the $60$ output terrain maps. The composition of roughness for each DEM is analyzed. We categorize each DEM pixel as having low, semi, or high roughness, as described in Sec.~\ref{sec:dem_filt_rand_sel}. The terrain covered by each roughness category is calculated and presented as an area percentage. To avoid bias, we calculate these percentages before applying the morphological filter. We also do not include the five-meter border in the DEM analysis since it will not be traversed. The areas of low, semi, and high roughness in each terrain are grouped by the $D$ value used in the high-frequency DEM for respective terrain generation. The terrain roughness results are presented in Sec.~\ref{sec:terrain_analysis}, using bar graphs to show the correlation between $D$ and the median area of each roughness category.

We then study how terrain traversability for the AGV is correlated with $D$. The AGV traverses $20$ random, straight-line paths in each of the $60$ terrain maps. The AGV's success rate is used as a metric to measure the traversability of the terrain maps. We define success rate as the percentage of successful missions traversed from start to goal in each map, where
\begin{equation}
success\ rate = (N_{success}/N_{trials})*100.
\label{eq_SR}
\end{equation}
In Eq.~\ref{eq_SR}, $N_{success}$ is the number of successful A-to-B traversals in a terrain. The total number of A-to-B trials, equal to $20$ in our experiments, is shown as $N_{trials}$.

Due to the rough terrain, the AGV may not complete its mission due to tipping over or getting stuck while en route. To detect failures, we monitor the AGV's orientation and location data throughout missions. If the pitch or roll of the AGV exceeds plus or minus 75\textdegree, we consider it to have tipped over. Similarly, if the AGV fails to move more than $0.2~m$ within $30~s$, we consider it to be stuck. In either case, the mission has failed and the simulation trial is ended.

We also analyze the RMS vertical acceleration and RMS pitch and roll rates only for successful missions. These dynamics serve as a measure of the AGV's response to different terrains encountered. We analyze successful missions to understand the impact $D$ has on AGV traversability that is not captured by success rates. Successful missions are also analyzed to remove potential bias in the vehicle dynamics experienced in failed missions such as the AGV tipping over. Vertical acceleration is calculated from the AGV's Z (elevation) time-series position data. We differentiate the position data twice to obtain vertical acceleration. Pitch and roll rates are determined by differentiating the AGV orientation time-series data. Pitch is measured relative to the AGV's transverse axis, while roll is measured relative to its longitudinal axis. Both pitch and roll measurements are referenced from the AGV in a level orientation. Traversal data is collected at $20~hz$. We use RMS metrics since we are primarily interested in comparing respective metric magnitudes, and not concerned with directions. Mission traversal times are also measured for successful missions. Traversal time is the time in seconds it takes the AGV to complete a straight-line path. Since the number of successful trials varies by the terrain map used, we aggregate the successful traversal metric results by $D$ used in terrain generation. Success rate and traversal performance metrics for successful trials are presented in Sec.~\ref{sec:results:trav}.

For our AGV traversal analysis, we present results in terms of the median and inner-quartile range (IQR) with boxplots. The boxplots show IQR, where 
\begin{equation}
IQR = Q_3 - Q_1,
\label{eq_IQR}
\end{equation}
such that $Q_3$ and $Q_1$ are the third and first quartile data values, respectively. The IQR measures the spread of the center $50\%$ of a respective metric's data points. We use median and IQR metrics since our data tends to be skewed and not normally distributed about its mean. 

\section{Results and Discussion} \label{sec:results}
We present our results here and discuss the effects of the fractal dimension on terrain roughness and AGV straight-line path traversals. Fractal dimensions of $2.3$, $2.45$, and $2.6$ are each used to generate $20$ unique DEMs. We first analyze the DEM roughness composition statistics of low, semi, and high roughness, described in Sec.~\ref{sec:dem_filt_rand_sel}. Success rates of the AGV traversing $20$ random missions in each of the $60$ terrains in UE are analyzed. The RMS vertical accelerations, RMS pitch and roll rates, and traversal times are analyzed only for successful trials in our $1,200$ trial study.  

\subsection{Terrain Roughness Analysis}
\label{sec:terrain_analysis}
\begin{figure}[h!]
\centering\includegraphics[width=0.5\textwidth]{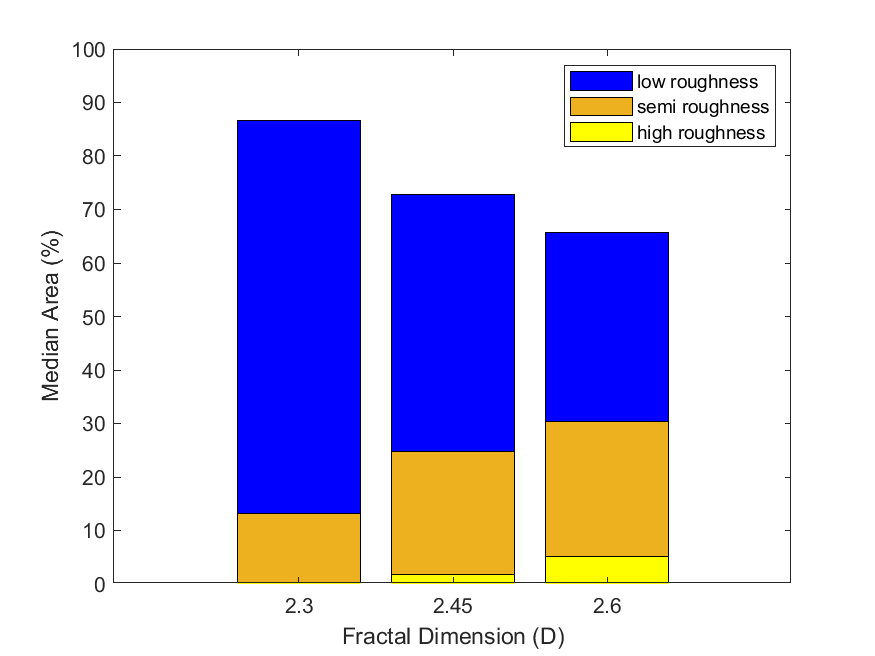}
\caption{A bar graph of the multifractal digital elevation maps' (DEM) median areas of low (a), semi (b), and high roughness (c), all by the fractal dimension used in the high-frequency DEM. Twenty terrain maps are generated for each of the three fractal dimension values of 2.3, 2.45, and 2.6, for a total of 60 maps.}
\label{fig:cdp}
\end{figure}

The roughness of the generated DEMs is analyzed by the fractal dimension used in the high-frequency DEM of our terrain generation process. Figure~\ref{fig:cdp} presents the median terrain roughness for each roughness category versus the fractal dimension used. The results of our terrain generation experiments show that as $D$ increases from $2.3$ to $2.45$ and from $2.45$ to $2.6$, the median area of low-roughness terrain decreases $13.8\%$ and $7.16\%$, respectively. On the other hand, the median area of semi-rough terrain increases $11.7\%$ and $5.63\%$ as $D$ increases from $2.3$ to $2.45$, and $2.45$ to $2.6$, respectively. Additionally, the median area of high-roughness terrain increases $1.54\%$ and $3.33\%$ as $D$ increases from $2.3$ to $2.45$ and $2.45$ to $2.6$, respectively.


These results indicate that increasing $D$ in the high-frequency DEM used in our approach leads to rougher terrain on average. This is explained by the elevation assignments in our DEMs as we adjust $D$. Looking at Eq.\ref{eq_WM}, as we increase $D$, the amplitude of the summed sinusoids increases exponentially, creating a rougher surface. From Eq.~\ref{eq_A}, we see that as $D$ increases, the scaling variable $A$ is scaled down exponentially, creating a flatter surface. However, we re-scale the low-, mid-, and high-frequency DEMs near the same scale before combining them. This results in the high-frequency DEM contributing and controlling the roughness realized in the multifractal DEM. 

Our terrain generation method allows us to measure AGV performance over trials with varied surface characteristics. By testing AGVs in terrains generated with various combinations of fractal dimensions, we can avoid biased results and ensure the AGV can handle changes in terrain difficulty. Our terrain generation also enables us to assess AGV performance in terrains inaccessible for full-vehicle field tests. This is a crucial step in ensuring the AGV is robust enough to perform efficiently in different scenarios. In the next section, we present the results of the AGV straight-line path traversals to understand how $D$ affects traversability.

\subsection{AGV Traversability Analysis}
\label{sec:results:trav}

\begin{figure}[h!]
\centering\includegraphics[width=0.5\textwidth]{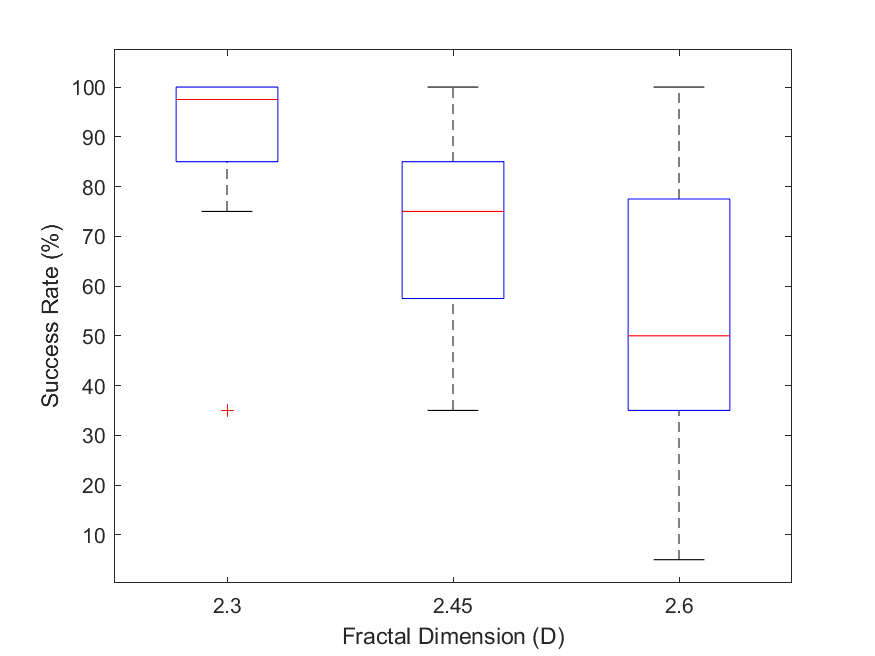}
\caption{Boxplots of the autonomous ground vehicle's (AGV's) success rate statistics in all terrains. The boxplots are grouped by the fractal dimension used in the high-frequency digital elevation map for generation. Twenty randomly selected A-to-B straight-line paths are traversed by the AGV in 60 terrain maps, totaling 1,200 trials. The success rate is calculated for each terrain map.}
\label{fig:SR_boxplot_all}
\end{figure}

The success rate of the AGV in each terrain map is calculated based on the $20$ straight paths attempted. The success rate statistics across all terrain maps by the fractal dimension used are displayed in Fig~\ref{fig:SR_boxplot_all}. The vertical axis shows the AGV success rate in each terrain. The horizontal axis shows the corresponding $D$ value used in the high-frequency DEM for terrain generation. The red horizontal line in each boxplot indicates the median roughness composition in each set of $20$ terrains. The lower and upper box limits represent the $25\%$ ($Q_1$) and $75\%$ ($Q_3$) percentiles, respectively. Vertical dotted lines extend to the minimum and maximum terrain compositions, excluding outliers. Outliers for all boxplots are indicated as a red "$+$" if an upper bound 
\begin{equation}
\label{eq_outliers1}
B_{upper} = (Q_3 + 1.5*IQR)
\end{equation}
or lower bound
\begin{equation}
\label{eq_outliers2}
B_{lower} = (Q_1 - 1.5*IQR)
\end{equation}
is exceeded. Here we see that the median success rate across terrains decreases $22.5\%$ as $D$ increases from $2.3$ to $2.45$, and decreases $25\%$ as $D$ increases from $2.45$ to $2.6$.

The results of our traversal experiments show that the success rates of the AGV are affected by our control of terrain roughness. We observe that $D$ is negatively correlated with the median success rate. This is explained by the increase in the median semi- and high-roughness terrain areas and the decrease in the median low-roughness areas. This tradeoff in roughness increases the chances of the AGV encountering terrain that impedes mission completion. The results highlight the effect of greater quantities of rougher terrain on traversability. Controlling the terrain's roughness can aid in emulating outdoor rough terrains for AGV testing. The results also highlight the need for a terrain-aware path planner to increase the likelihood of mission success.

\begin{figure*}[hbt!]
\centering
\subfigure[]{\label{fig:z_accel_boxplot}\includegraphics[width=3.4in]{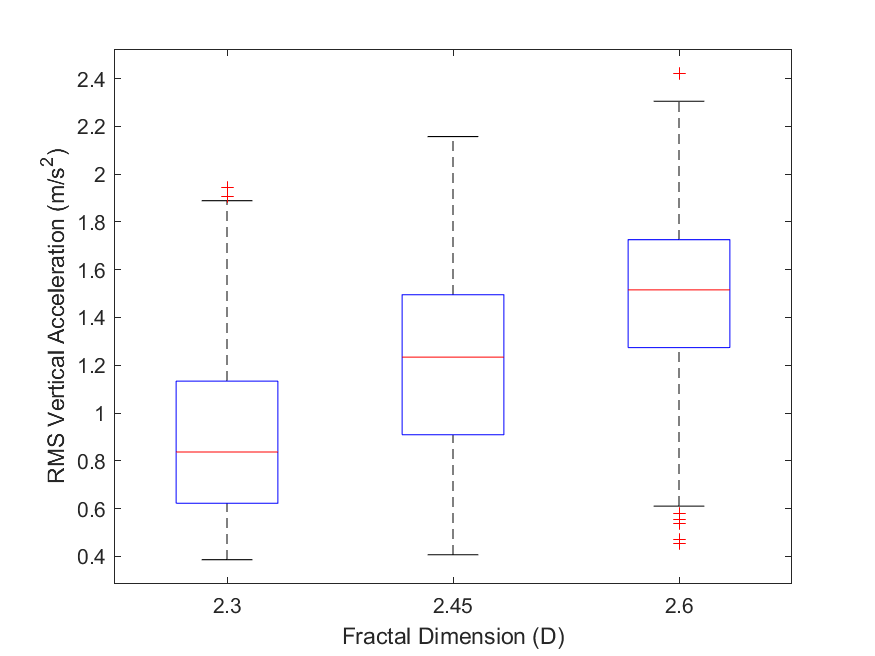}}
\subfigure[]{\label{fig:ptch_boxplot}\includegraphics[width=3.4in]{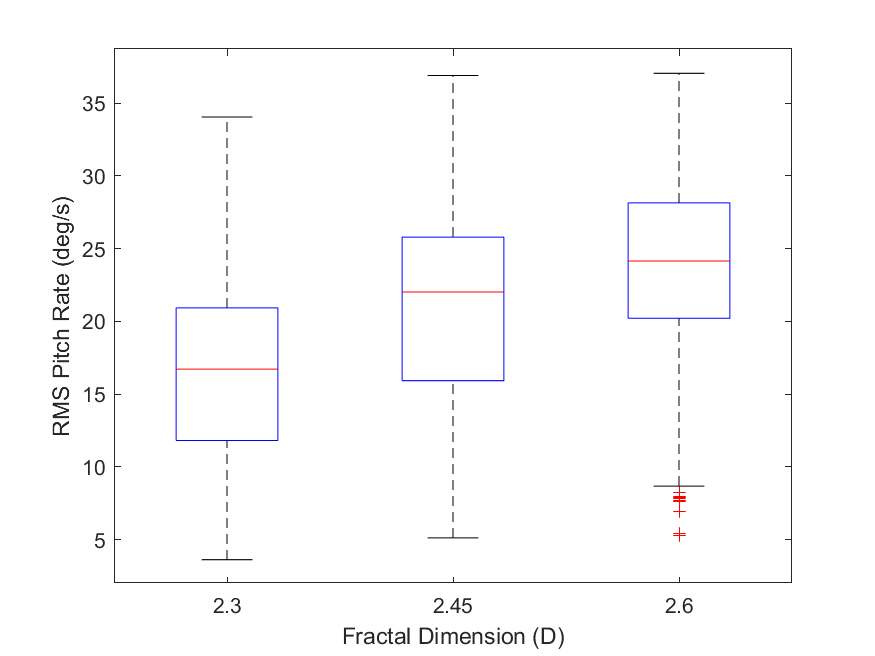}}
\subfigure[]{\label{fig:roll_boxplot}\includegraphics[width=3.4in]{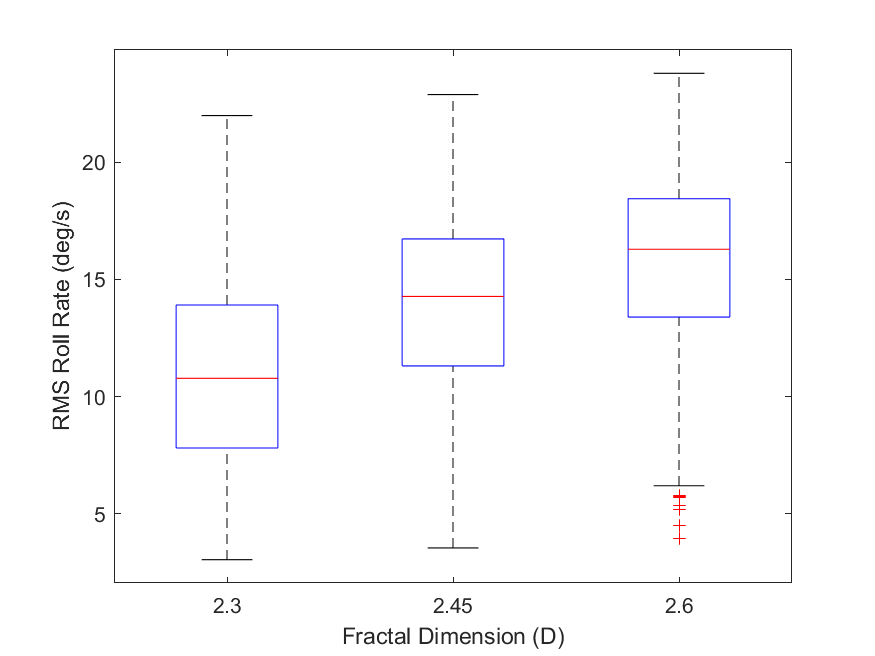}}
\subfigure[]{\label{fig:trav_time_boxplot}\includegraphics[width=3.4in]{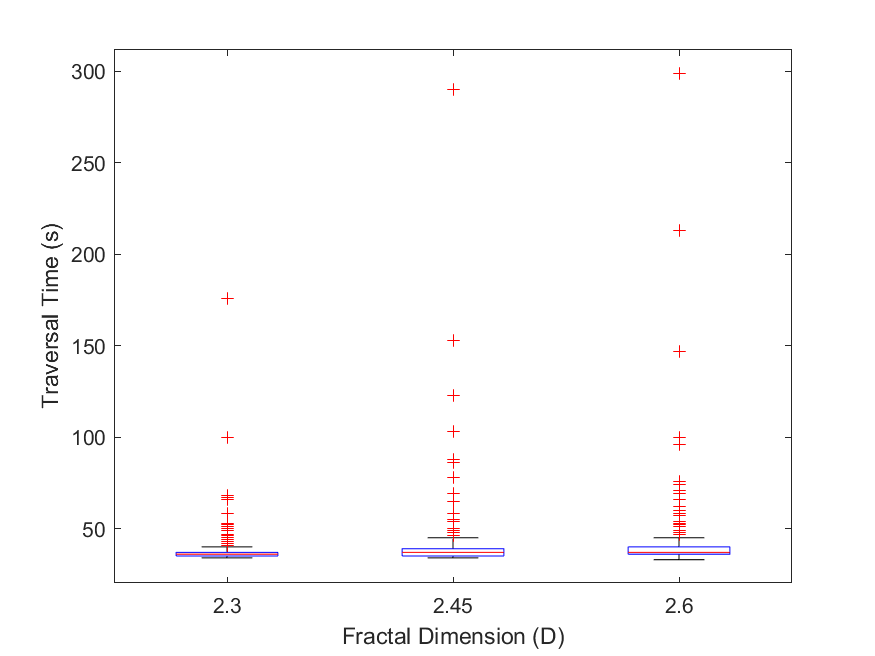}}
\caption{Boxplots of the autonomous ground vehicle's root-mean-squared (RMS) vertical accelerations (a), RMS pitch rates (b), RMS roll rates (c), and traversal times (d) for successful trials across all $60$ terrain maps by the digital elevation maps' fractal dimension. Note that the number of successful trials is $361$, $282$, and $225$ for all terrains generated with fractal dimensions of $2.3$, $2.45$, and $2.6$, respectively.}
\label{fig:success_traversal_boxplots}
\end{figure*}

The boxplots shown in Fig.~\ref{fig:success_traversal_boxplots} display the traversal metrics for successful missions. They depict the RMS vertical accelerations (\ref{fig:z_accel_boxplot}), RMS pitch and roll rates (\ref{fig:ptch_boxplot} - \ref{fig:roll_boxplot}), and traversal times (\ref{fig:trav_time_boxplot}) aggregated across all $20$ terrain maps of each $D$ group. By analyzing this data, we can observe the changes in vehicle response and traversal times as $D$ increases that are not captured by success rates. Note that of the $400$ AGV trials for each $D$ value of $2.3$, $2.45$, and $2.6$ used, there were $361$, $282$, and $225$ successful trials, respectively.

As $D$ increases from $2.3$ to $2.45$ and from $2.45$ to $2.6$ we find the following. The median RMS vertical acceleration increases incrementally $0.397~m/s^2$ ($47.4\%$) and $0.281~m/s^2$ ($22.8\%$), respectively. The median RMS pitch rate increases incrementally $5.29~deg/s$ ($31.7\%$) and $2.13~deg/s$ ($9.67\%$), respectively. The median RMS roll rate increases incrementally $3.5~deg/s$ ($32.4\%$) and $2.02~deg/s$ ($14.1\%$), respectively. The median traversal time only changed when $D$ increased from $2.3$ to $2.45$, with an increase of $1~s$ ($2.78\%$). Note that the number of outliers in the traversal time data is $37$, $19$, and $30$ for $D$ values of $2.3$, $2.45$, and $2.6$, respectively.

The increase in the median RMS vertical acceleration and median RMS pitch and roll rates suggests that even in successful trials, the AGV experiences rougher terrain on average as $D$ increases. This data agrees with the terrain roughness results in Sec.~\ref{sec:terrain_analysis} which shows an increase in the average areas of semi- and high-roughness terrain and a decrease in the average area of low-roughness terrain as $D$ increases. Looking at Fig.~\ref{fig:trav_time_boxplot}, it is not surprising that the median traversal times are relatively unaffected since we are analyzing only successful trials in missions defined with straight-line paths of the same Euclidean distance. However, this shows that the rougher terrain experienced by the AGV did not substantially affect median traversal times in successful missions. The traversal time outliers represent trials where the AGV's path was impeded more significantly by rough terrain. These outliers are cases where the AGV got temporarily stuck but continued to move such that a failure was not flagged. Another cause of the outliers is rough terrain causing the AGV to deviate from the straight paths, causing longer paths to be taken. 

The results of our experiments demonstrate that varying the fractal dimension in the high-frequency DEM affects the performance of the simulated AGV. Although the AGV still completes missions as $D$ increases, the average increase in its RMS vertical accelerations and pitch and roll rates show that the AGV's kinematics are challenged more as $D$ increases. By increasing $D$, our terrain generation approach can be used to stress test a path planner or navigation method to a degraded performance or failure. This can help identify algorithm weaknesses which can then be analyzed and corrected for when operating in dangerous real-world terrain.

Note that the results discussed in this paper are specific to a modeled Clearpath Husky AGV. However, the terrain generation method outlined in this work can be utilized for vehicles of similar size and characteristics. This method can also be applied to vehicles of different sizes with the proper consideration of terrain scaling.

\section{Conclusions and Future Work} \label{sec:conclusions}
In this paper, we presented a multifractal approach for generating artificial off-road terrains with varied roughness. Our method combines DEMs of low-, mid-, and high-frequency surface roughness, which are defined by different W-M function parameters. We tested if we could control terrain roughness by adjusting the fractal dimension in the high-frequency DEM used in our method. Additionally, we tested how the performance of a non-terrain-aware AGV in A-to-B traversals would correlate with changes in the fractal dimension.

We generated $60$ unique DEMs using three different fractal dimension values of $2.3$, $2.45$, and $2.6$. We computed a gradient map from each respective DEM to quantify the areas of low-, semi-, and high-roughness terrain. The DEMs were rendered into 3D terrains in UE for AGV testing.

We found that the median area of low-roughness terrain decreased $13.8\%$ and $7.16\%$, the median area of semi-rough terrain increased $11.7\%$ and $5.63\%$, and the median area of high-roughness terrain increased $1.54\%$ and $3.33\%$, all respectively as we increased the fractal dimension from $2.3$ to $2.45$, then to $2.6$. The results show our ability to control roughness with our approach.

To test terrain traversability with a modeled Clearpath Husky AGV, we conducted $20$ randomized straight-line path traversals in each terrain. We found that the median success rate of the vehicle decreased $22.5\%$ and $25\%$, respectively as the fractal dimension increased incrementally from $2.3$ to $2.45$, and from $2.45$ to $2.6$. Successful trials show that the median RMS vertical accelerations, median RMS pitch and roll rates, and median traversal times also increase with $D$. The traversal metrics show that the AGV encountered rougher terrain and was more kinematically challenged on average as the fractal dimension increased.

Our method allows for control of the roughness in randomly generated terrain surfaces using the 3D W-M function, extending on prior work that generated terrain profiles. The terrain output has varied roughness such that path planning and navigation are non-trivial. The DEMs generated can also be easily adapted to other simulation environments. The proposed method is practical for creating off-road terrain with varying levels of difficulty. Control of terrain roughness is useful for selecting terrain for AGV evaluations in simulation. In future work, we will use this terrain generation approach in evaluating a terrain-aware AGV path planning and navigation method.

\section*{Acknowledgment} 
The authors wish to acknowledge the financial support provided by the USD/R\&E (The Under Secretary of Defense-Research and Engineering), National Defense Education Program (NDEP)/BA-1, Basic Research in the Science Mathematics and Research for Transformation (SMART) Scholarship Program. These views and/or findings do not represent the views of the SMART Program, Department of Defense, or United States Government. (Corresponding author: Casey Majhor.) We would also like to acknowledge contributions by Sam Kysar, Parker Young, Zach Jeffries, Ian Mattson, Jake Carter, Chaz Cornwall, and Logan Schexnaydre.



\section*{Conflict of Interest}
There are no conflicts of interest.

\section*{Data Availability Statement}
The datasets generated and supporting the findings of this article
are obtainable from the corresponding author upon reasonable
request.

%
% ...

\bibliographystyle{asmems4}

\bibliography{asme2e}

\end{document}